\documentclass{sig-alternate-05-2015}

   \usepackage{graphicx}
  \graphicspath{{../pdf/}{../jpeg/}}
  \DeclareGraphicsExtensions{.pdf,.jpeg,.png}
  
  \usepackage[export]{adjustbox}
  \usepackage{caption}
	\usepackage{subcaption}
    \usepackage{etoolbox}
\makeatletter
\patchcmd{\maketitle}{\@copyrightspace}{}{}{}
\makeatother
\begin{document}





\conferenceinfo{PLDI '13}{June 16--19, 2013, Seattle, WA, USA}

%

\title{Design, Analysis \& Prototyping of a Semi-Automated Staircase-Climbing Rehabilitation Robot}
%
%
%
%
%

\numberofauthors{9} 
%
\author{
%
%
\alignauthor
Siddharth Jha\\
       \affaddr{IIT Kharagpur}
       \email{thesidjway@iitkgp.ac.in}
\alignauthor
Himanshu Chaudhary\\
       \affaddr{IIT Kharagpur}\\
       \email{himanshuchaudhary} \titlenote{@iitkgp.ac.in}
\alignauthor Swapnil Satardey\\
       \affaddr{IIT Kharagpur}\\
       \email{swapnilsatardey@iitkgp.ac.in}
\and 
\alignauthor Piyush Kumar\\
       \affaddr{IIT Kharagpur}\\
       \email{piyushomega@gmail.com}
\alignauthor Ankush Roy\\
       \affaddr{IIT Kharagpur}\\
       \email{ankushroy@iitkgp.ac.in}
\alignauthor Aditya Deshmukh\\
       \affaddr{IIT Kharagpur}\\
       \email{adityadeshmukh281997@iitkgp.ac.in}
}
\additionalauthors{Gopabandhu Hota (IIT Kharagpur,
{\texttt{gopabandhuhota1@iitkgp.ac.in}})\\ Dishank Bansal (IIT Kharagpur,
{\texttt{dishank.bansal@iitkgp.ac.in}})\\ Saurabh Mirani (IIT Kharagpur,
{\texttt{mirani.saurabh@iitkgp.ac.in}})\\}
\date{06 November 2017}

\maketitle 
\begin{abstract}
In this paper, we describe the mechanical design, system overview, integration and control techniques associated with SKALA, a unique large-sized robot for carrying person with physical disabilities, up and down staircases. As a regular wheelchair is unable to perform such a maneuver, the system functions as a non-conventional wheelchair with several intelligent features. We describe the unique mechanical design and the design choices associated with it. We showcase the embedded control architecture that allows for several different modes of teleoperation, all of which have been described in detail. We further investigate the architecture associated with the autonomous operation of the system.
\end{abstract}

%
%

%
%

%
%



\section{Introduction}
A regular wheelchair is designed to move only on plain ground and being mechanically driven, run at a very low speed. Although powered wheelchairs are available, and mobility scooters are getting common, all of these fail to address a very basic job of traversing up and down stairs reliably that limits their functionality to flat ground. A lot of work has been done on negotiating staircases by robots with people seated. This has been conceptualized in \cite{lawn2000towards}. A breakthrough in the design of staircase traversal using conventional wheelchairs has been shown in \cite{morales2006kinematic}  and \cite{morales2006coordinated}. The works show the use of two decoupled mechanisms for stair negotiating and slope adjustments. With the rise in computing power and reduction in size of computers, autonomous and semi-autonomous wheelchairs have been proposed as well. The problem with all these rehabilitation systems arises when a subject suffering from paraplegia, which is a medical condition involving impairment of lower extremities, is seated. Users suffering from this condition may lack sufficient strength to lift themselves. All these discussed solutions lack the flexibility of the subject lifting him/herself in an upright or any variable seating orientation. 

The objective of this research is to develop a low-cost yet reliable mobile platform which is capable of seating a subject and is equipped with sufficient connectivity, flexibility and compute to test various control algorithms, teleoperation methods and automation techniques. The platform should enhance the usage scenario of a regular powered wheelchair with its staircase traversing capability. It should provide flexibility to the user to lift him/herself in a desired seating orientation. Also, the system developed should be rugged, have good scope of expansion in both mechanical and embedded design, should be capable of running at good speeds and should not be limited to a particular staircase design.

This paper is divided into six total sections. We start off with the mechanical design aspects, and demonstrate the design choices for the system. Next we demonstrate the calculations \& simulations performed for the choice of actuators. We also show the Size Weight and Power (SWAP) design choices along with details of fabrication. In the following sections, we describe the power flow, embedded and automation architecture. Finally we conclude with the prototyping results.

\section{Mechanical Design}
 
 The mechanical design primarily addresses the problem of stair-climbing and truncating the hassle of locomotion for a paraplegic person. Optimal user flexibility and maneuverability, which conventional devices lack, are taken as the prime motives. Furthermore, it facilitates convenient motion on horizontal surfaces by the control of two front wheels, which are directly connected to two motors, and two caster wheels, which assist in turning left or right. Caterpillar tracks are observed to be the most convenient solution as they enable smooth and continuous motion on the stairs \cite{02studyof}. They are indirectly driven by the two motors using a chain and sprocket mechanism. The loading base plate is connected to two actuators which help in keeping the plate level horizontal during the climb. The upper chassis houses the user who is strapped into position by a cushion and a harness.

The study of the mechanical design has been divided into 2 parts, namely  User support assembly and Caterpillar track.

\subsection{User Support Assembly}
It consists of a T-shaped vertical rod welded to the loading base plate. The user is held in mid-air with the help of a cantilever arm which is hinged to the vertical rod and can lift him/herself to desired seating orientation using a linear actuator. Actuator force varies with angle $\theta$ made by the cantilever arm with the horizontal axis as
\begin{equation}
\label {eq:actuatorforce}
F= \frac{(a+b) \times mg \times cos(\theta)}{sin(\gamma)}
\end{equation}

\begin{figure}[!h]
\centering	
\captionsetup{justification=centering}
\begin{minipage}[t]{.4\linewidth}
	\centering
  \includegraphics[width=0.8\linewidth]{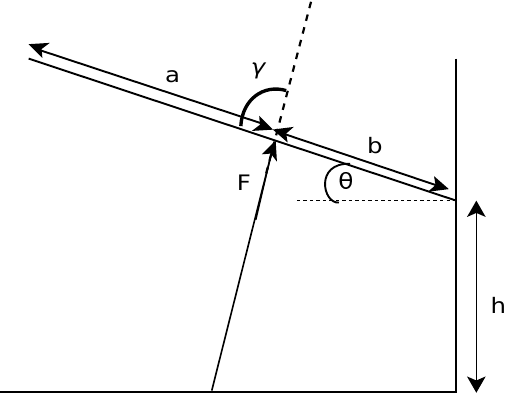}
  \captionof{figure}{Line Diagram showing the parameters}
  \label{fig:LineDiagram}
\end{minipage}%
\begin{minipage}[t]{.6\linewidth}
	\centering
 \includegraphics[width=1.0\linewidth]{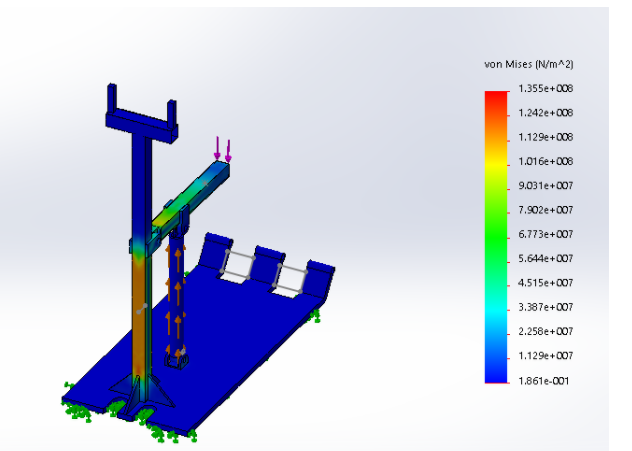}
  \captionof{figure}{Stress Profile}
  \label{fig:stress}
\end{minipage}
\end{figure}

The relation between the angles formed between by the actuators and cantilever arm (as shown in the Fig (\ref{fig:LineDiagram})) is
\begin{equation}
\label {eq:angle}
sin(\theta + \gamma - 90^{o})= \frac{2b \times sin(\frac{\theta}{2}) \times cos(\gamma + \frac{\theta}{2})}{h}
\end{equation}

where a, b and h are the lengths shown in Fig 1.Assuming a person of average height (177cm), values of a, b and h were calculated to be 33.5cm, 22.5cm and 60cm respectively.

Following are the plots which shows the variation of actuator forces with $\theta$ :

\begin{figure}[!h]
\centering	
  \includegraphics[width = \linewidth]{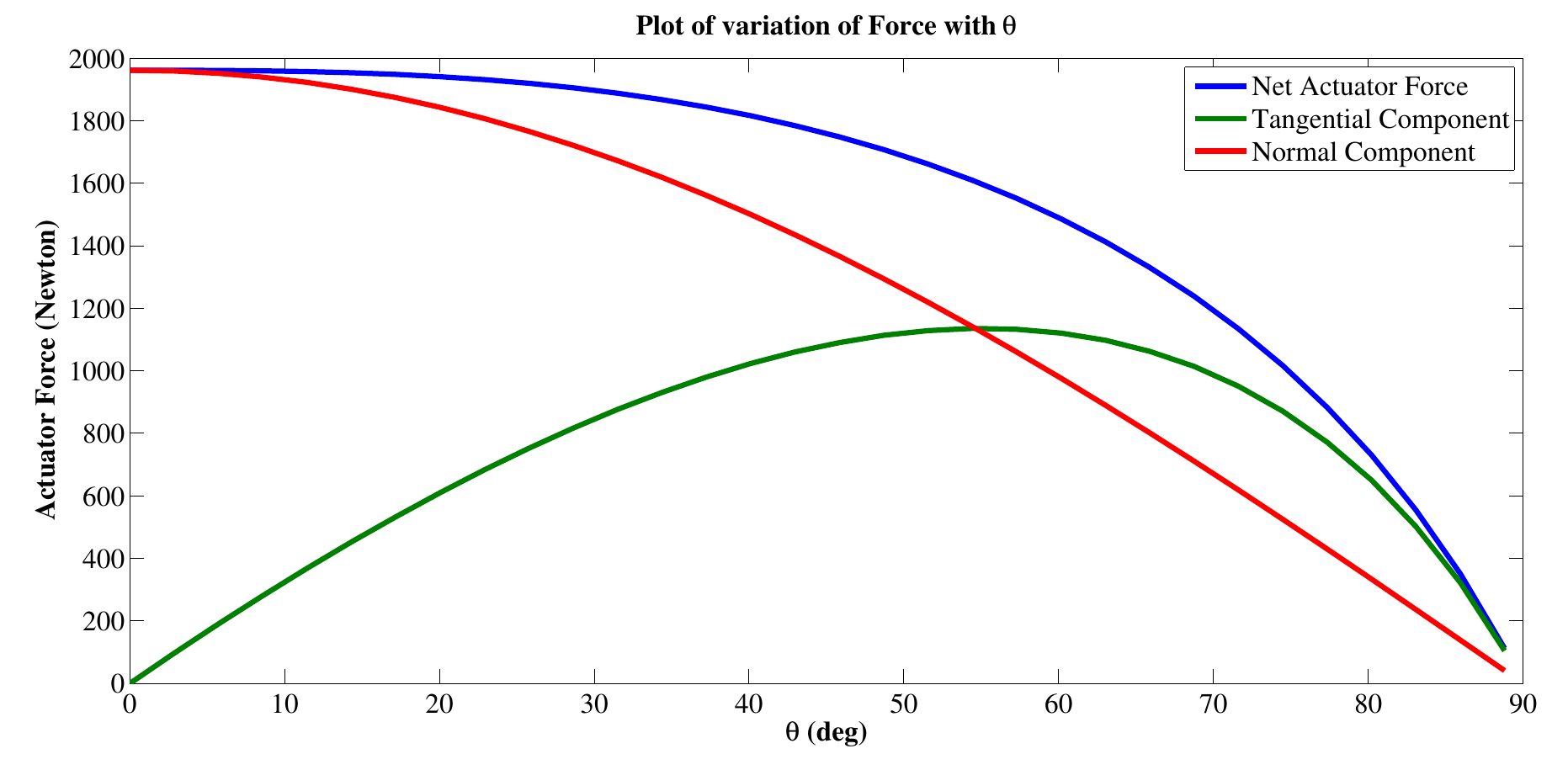}
  \caption{Variation of Force with $\theta$}
  \label{fig:force}
\end{figure}

The robot is designed to bear the load of a user weighing 120kg. The corresponding stress profile of the model when subjected to  maximum load, calculated from (\ref{eq:actuatorforce}) is shown in Fig \ref{fig:stress}. The max. shear force is 1130 N at the actuator hinge, which is well below the fracture strength of the weld. 
 
 The user support assembly is hinged to the caterpillar track and is supported by two actuators in the front, capable of  lifting up the assembly during the stair climbing operation and hence, keeps the platform horizontal. The whole user support assembly is made of cast iron.

\subsection{Caterpillar Track}
The stair-climbing mechanism is based on caterpillar drive. This mechanism involves a significant amount of design considerations and analysis to ensure safe movement on stairs.

\subsubsection{Design Overview}
The caterpillar drive is chosen over other stair-climbing drives in accordance with the analysis in \cite{02studyof}.Track-based mechanisms are suitable for all kind of stairs and have a simple operation. 
The caterpillar drive is achieved through a double-sided timing belt stretched across three timing pulleys and supported by rollers(diameter- 5cm). This is done to ensure there is no slacking of the belt while on the stairs. A chain-sprocket type transmission with a gear-reduction ratio of 2:1 transfers the power from the motor to the track mechanism. Thus, the wheels and the tracks are powered simultaneously.

\begin{figure}[!h]
\centering
\captionsetup{justification=centering}
\begin{minipage}[t]{.5\linewidth}
	\centering
  \includegraphics[width=0.9\linewidth]{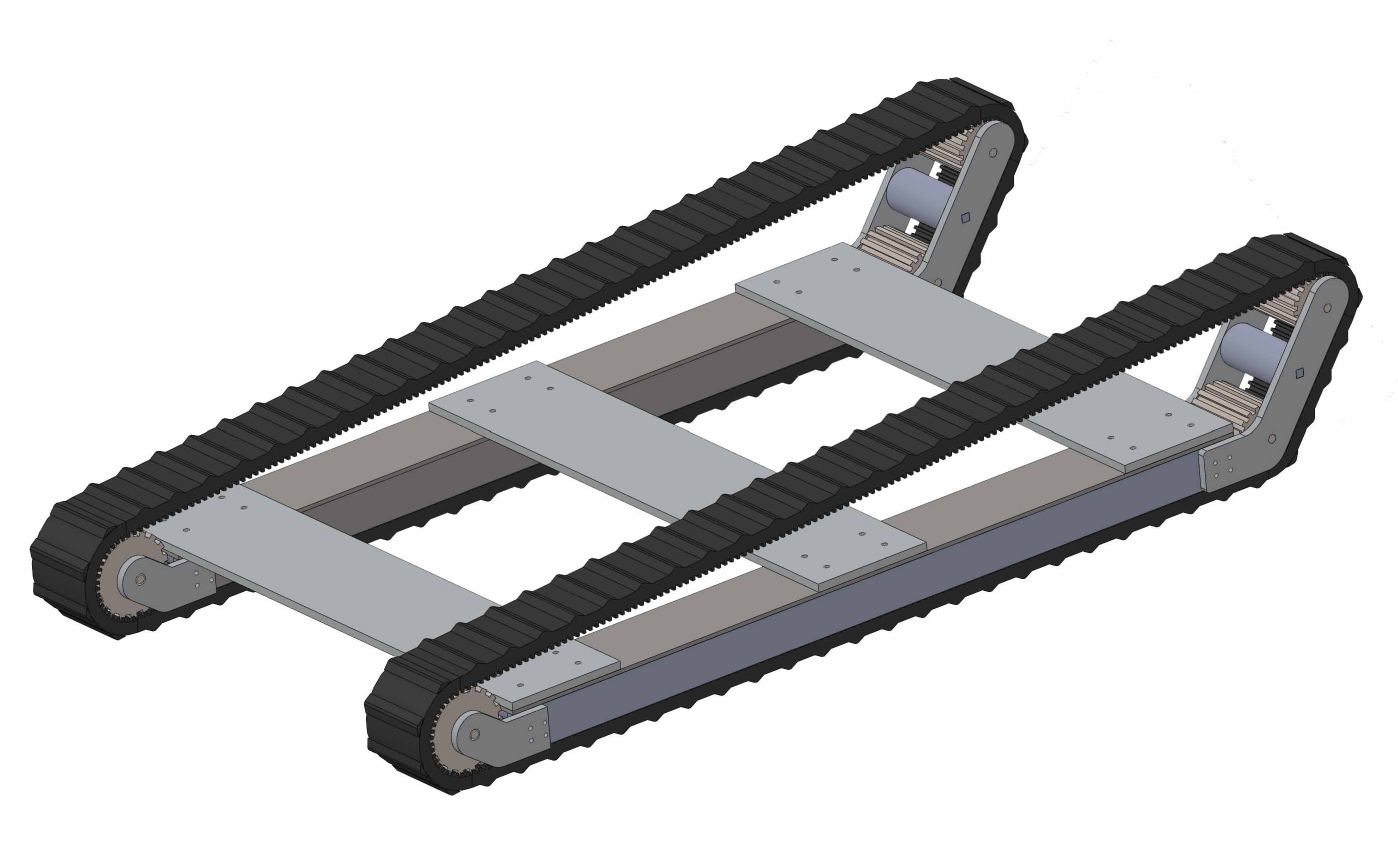}
  \captionof{figure}{CAD model of caterpillar drive mechanism}
  \label{fig:track}
\end{minipage}%
\begin{minipage}[t]{.5\linewidth}
	\centering
 \includegraphics[width=0.9\linewidth]{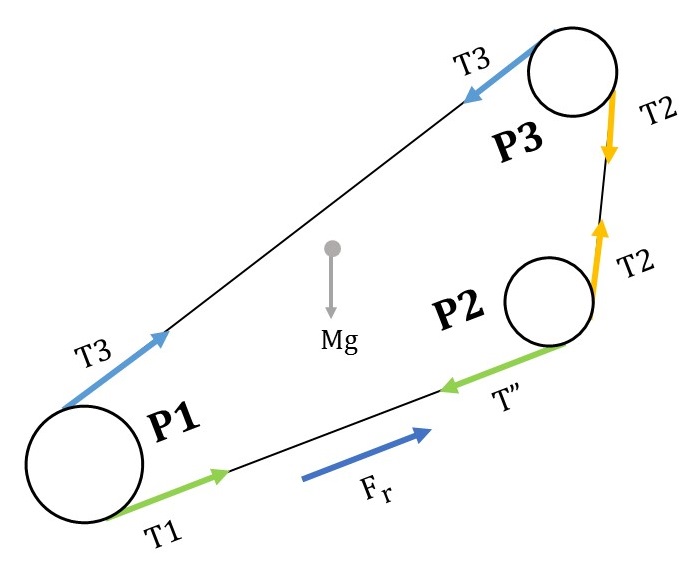}
  \captionof{figure}{Free Body Diagram of the robot when climbing up stairs of inclination $\theta$}
  \label{fig:fbd}
\end{minipage}
 
\end{figure}

\subsubsection{Track Components}
The belt used for the track is T20DL double sided brecoflex timing belts. This is a Polyurethane, steel-reinforced and trapezoidal waveform belt. The module is 4mm and pitch is 12.5mm. The custom-made timing pulleys are based on external spur gear design with a module of 4mm. Two of them have a radius of 3.6cm and the third one has a radius of 5cm. The pulleys are made of mild steel.
Rollers are made of Ultra-high-molecular-weight polyethylene (UHMW). It has a high impact strength and a greater abrasion resistance than carbon steel. It also has a low coefficient of friction and is light with density one-eighth of steel.

\subsubsection{Stair climbing dynamics \& motor specifications}
\begin{figure}[!h]
\centering	
  \includegraphics[width=0.8\linewidth]{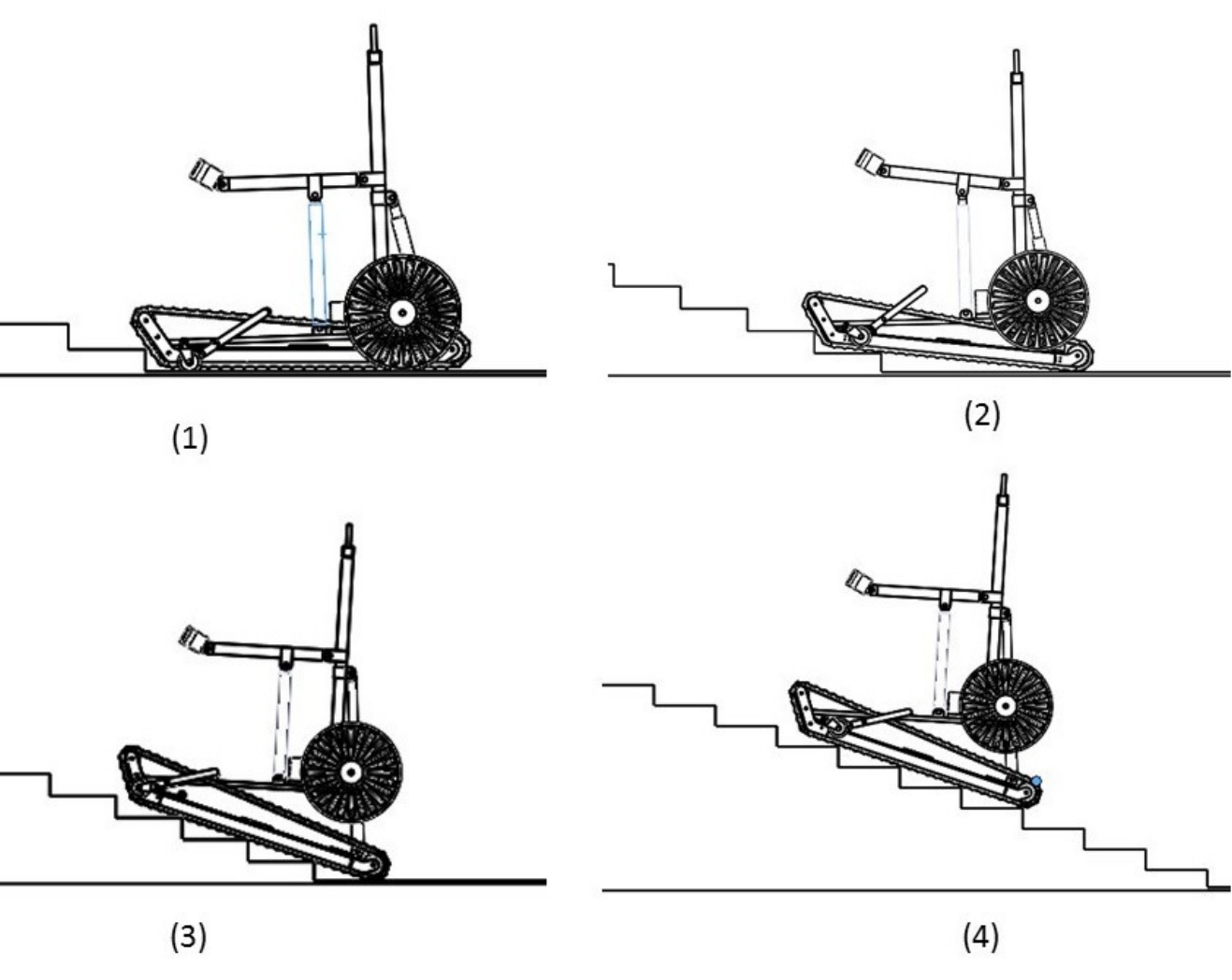}
  \caption{The stair climbing sequence}
  \label{fig:climb}
\end{figure}

Keeping the safety of the person into consideration, the robot has been designed to move at a maximum acceleration of 0.5 m/s$^{2}$ and negotiate stair angles up to 40$^\circ$. This angle limit also determines the stroke length of the front actuators, used to lift the base platform during stair climbing.
Since the output sprocket can be attached to any of the three timing pulleys, there are three possibilities of powering the caterpillar drive. Consider the robot is ascending the stairs with an acceleration 'a' and there is no slipping between the track and the stairs. From the analysis done in \cite{doi:stair}, one can infer that maximum torque is required when the robot is completely on the stairs. So, only the case when the robot is climbing up the staircase is considered.
The optimum position to power the track depends on two factors, the required torque to ascend the stairs \& the distribution of tensions throughout the belt.

\subsubsection*{Minimum required Torque (see Figure \ref{fig:fbd})}
To find the min. required torque, three cases (\ref{eq:P1}, \ref{eq:P2}, \ref{eq:P3}) of the relation between torque, acceleration and stair angle are considered, after powering pulley P1, P2 or P3 respectively.
\begin{equation}
\label{eq:P1}
\tau=R \times [(M+m-(m_{1}/2)) \times a+Mg \times sin(\theta)]
\end{equation}
\begin{equation}
\label{eq:P2}
\tau=r \times [(M+m_{1}) \times a+ Mg \times sin(\theta)]
\end{equation}
\begin{equation}
\label{eq:P3}
\tau=r \times [(M+m_{1}) \times a+Mg \times sin(\theta)]
\end{equation}

\begin{figure}[!h]
\centering	
  \includegraphics[width = \linewidth]{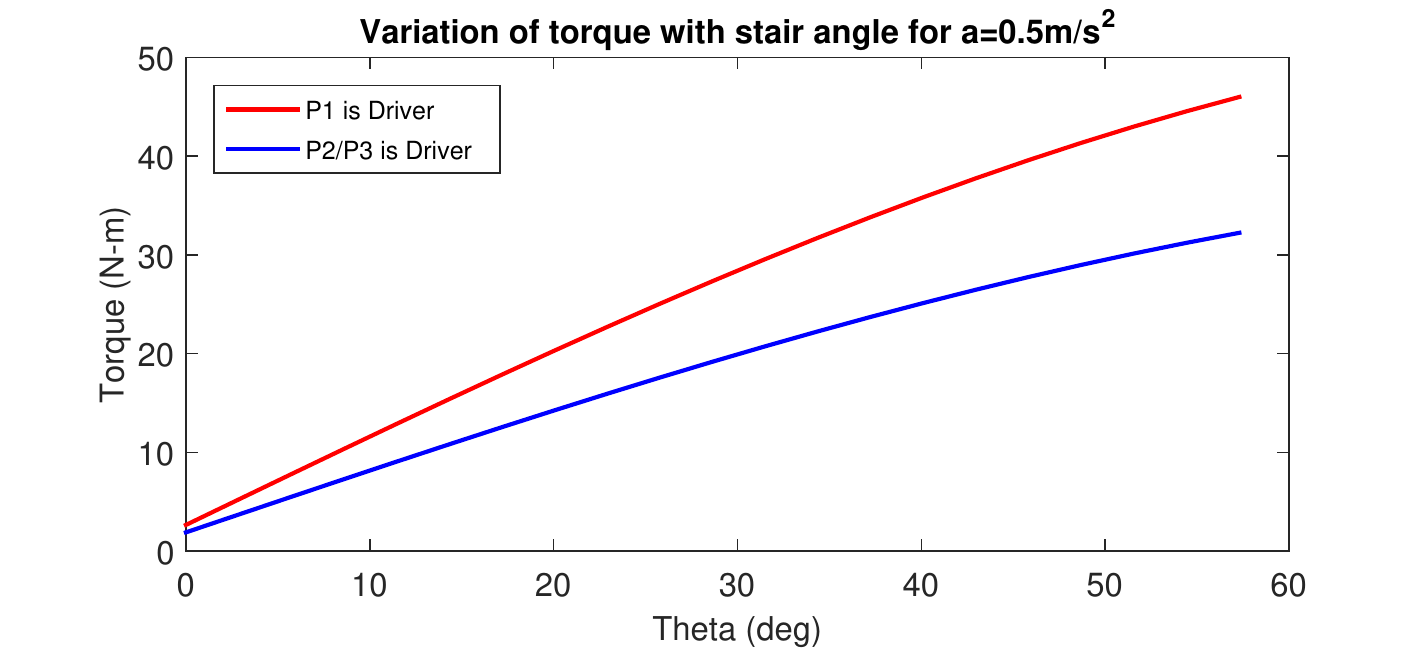}
  \caption{Variation of torque with stair angle for a=0.5 m/s$^{2}$}
  \label{fig:tau}
\end{figure}

\begin{figure}[!h]
\centering	
  \includegraphics[width=1.0\linewidth]{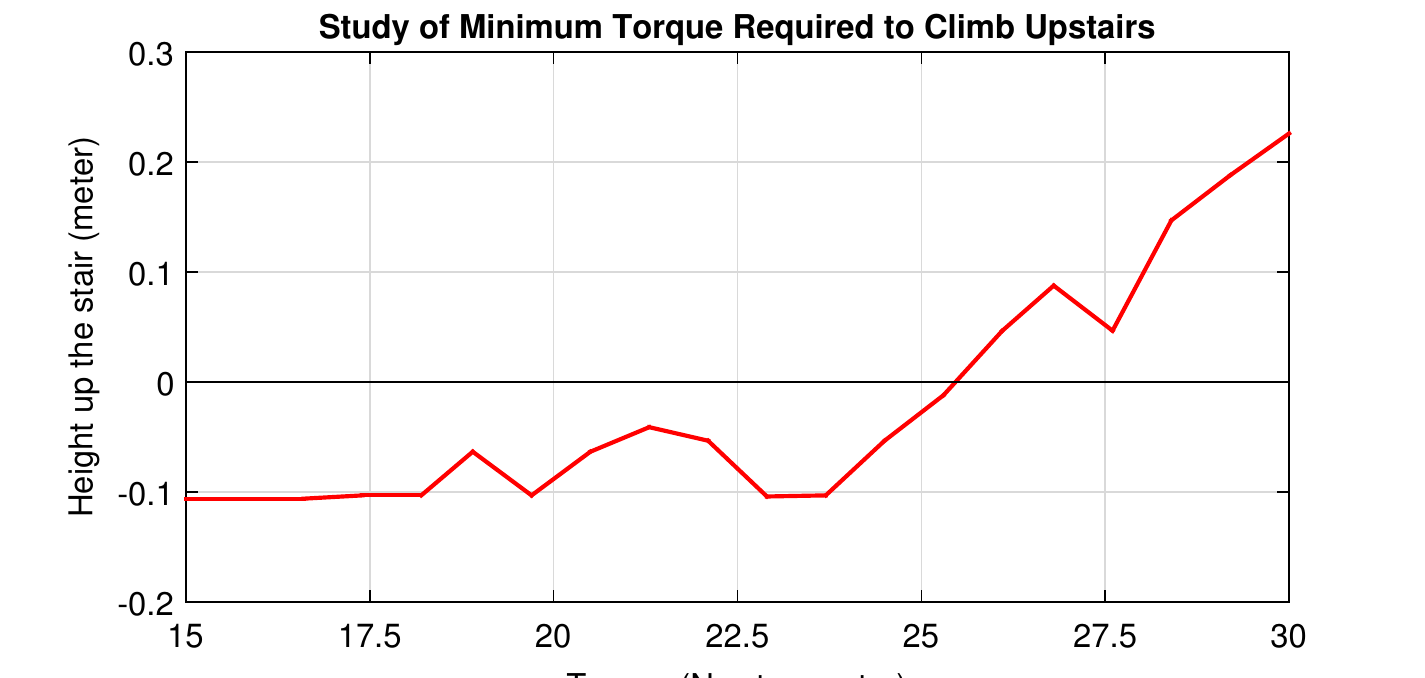}
  \caption{Simulation to find out minimum torque required to climb up the stairs (time=10sec). The point where the red line intersects y=0 line gives the required torque}
  \label{fig:simulation}
\end{figure}

Where $M$ is the mass of the robot, $m_{1}$ is mass of pulley P1, $m$ mass of pulleys P2 and P3, R is the radius of P1, r is radius of P2 \& P3 and $\theta$ is the stair angle. As Equations \ref{eq:P2} \& \ref{eq:P3} are the same, only cases P1 and P3 are considered. On plotting equation \ref{eq:P1} and \ref{eq:P3}, it is seen that there is no significant difference in torques if radius of the pulleys is same but if we decrease the radius then the required torque to ascend decreases significantly. 
Fig. \ref{fig:tau} illustrates the required torque to ascend stairs of $40^\circ$ inclination, when powering pulley P1 (R=5cm) and pulley P3 (r=3.6cm). So, in Case 1 the required torque is 35.8 N-m, while in case 2/3 the torque is significantly reduced to 25 N-m. Considering the case of powering P3 and climbing stairs of inclination of $40^\circ$, at a constant speed (a=0) gives the value of minimum required torque which is 22 N-m at each track. 
From the simulation shown in Fig. \ref{fig:simulation}, we get the required torque to be 25.4 N-m.

Since increasing the climb speed is not a major concern for stair climbing, the pulley radius has not been determined by wheel radius optimization, and rather the no interference condition and contact ratio for gears as given in \cite{theory} has been used to find out the minimum radius possible.

Formula for minimum no of teeth of pinion for no interference for rack and pinion is given as: 
\begin{equation}
N=\frac{2 \times f}{sin^{2}\alpha}
\label{eq:teeth}
\end{equation}

Contact Ratio is defined as:
\begin{equation}
m_{c}=\frac{N}{2\pi r_{b}}\times (\frac{a}{sin\alpha} + \sqrt[]{r_{o}^2-r_{b}^2} - r_{b}\times tan\alpha)
\label{eq:cr}
\end{equation}

Using equation \ref{eq:teeth}, N=18($\alpha$ (pressure angle)=$20^\circ $, f=1 (for addendum = module)).
Taking a standard module of 4mm, the diameter is determined to be 72mm and contact ratio is 1.75 (from equation \ref{eq:teeth}, \ref{eq:cr} and the relation $m=d_{p}/N$).
Thus, the minimum required radius of pulley for good operation of track is 3.6cm. So pulley P2 and P3 have radius of 3.6cm. The pulley P1 has a radius of 5cm to increase the mechanical advantage at the rear end while climbing upstairs.
$
$\subsubsection*{Distribution of Tension}
Fig. \ref{fig:fbd} shows the tensions in different parts of the belt. The magnitude of tensions will be in a particular order depending upon the driver pulley. The exact values of tension can be found out using equations (\ref{eq:P1}), (\ref{eq:P2}), (\ref{eq:P3}), length constraint equation \& stress-strain relationships. So, the order of the tensions are (Taking clockwise rotation of pulley gives forward motion):
\begin{enumerate}
\item Case 1: P1 is driver - $T1>T2>T3$
\item Case 2: P2 is driver - $T2>T3>T1$
\item Case 3: P3 is driver - $T3>T1>T2$
\end{enumerate}
So, for better grip on the stairs, the belt part in contact with the stairs should be taut. Hence, T1 should have the highest magnitude followed by T2 as during the start of ascend that part should be taut. So, Case 1 is preferable, but Case 3 has been opted because-
It is preferred to climb the robots in backward direction as it is much safer position in the event of a fall. So, base loading plate is hinged at P3. Unlike the case with P1, powering P3 gives the advantage of using only one motor for both ground locomotion and stair climbing. Thus, it is not possible to power P1 without the use of separate motors.

To tackle the loss in tension a tensioner is incorporated.
Taking into consideration all the factors a DC Motor (320W, 22Nm and 143-rpm), has been used to power each of the tracks. The motor is connected to pulley P3 through a chain-sprocket transmission with a gear reduction ratio of 2:1.

\section{Power Flow \& Embedded System Design}
\subsection{Power Flow Design}
The power system of SKALA consist of 2 lithium-polymer batteries (11.1V, 2700mAh, 25C each), 2 lead-acid batteries (12V, 26000 mAh each) and a power distribution unit. The 2 lithium-polymer batteries are used to power 3 linear actuator drivers rated at 16A and the two lead-acid batteries in series providing 24V total are connected to the power distribution board. The board routes power to the main motor drivers and contains a buck-converter-regulator circuit
to step down the 24V voltage to regulated 5V DC output to supply power to the low-voltage low-current sensors \& microcontrollers. Each of the motors are driven by the individual Cytron MDS40B motor driver which is controlled via input PWM signal from the microcontrollers and can supply an average load current of 40A (peak- 80A) to the motors.


\subsection{Embedded System Design}
\begin{figure}[!h]
\centering	
  \includegraphics[width = 0.9\linewidth]{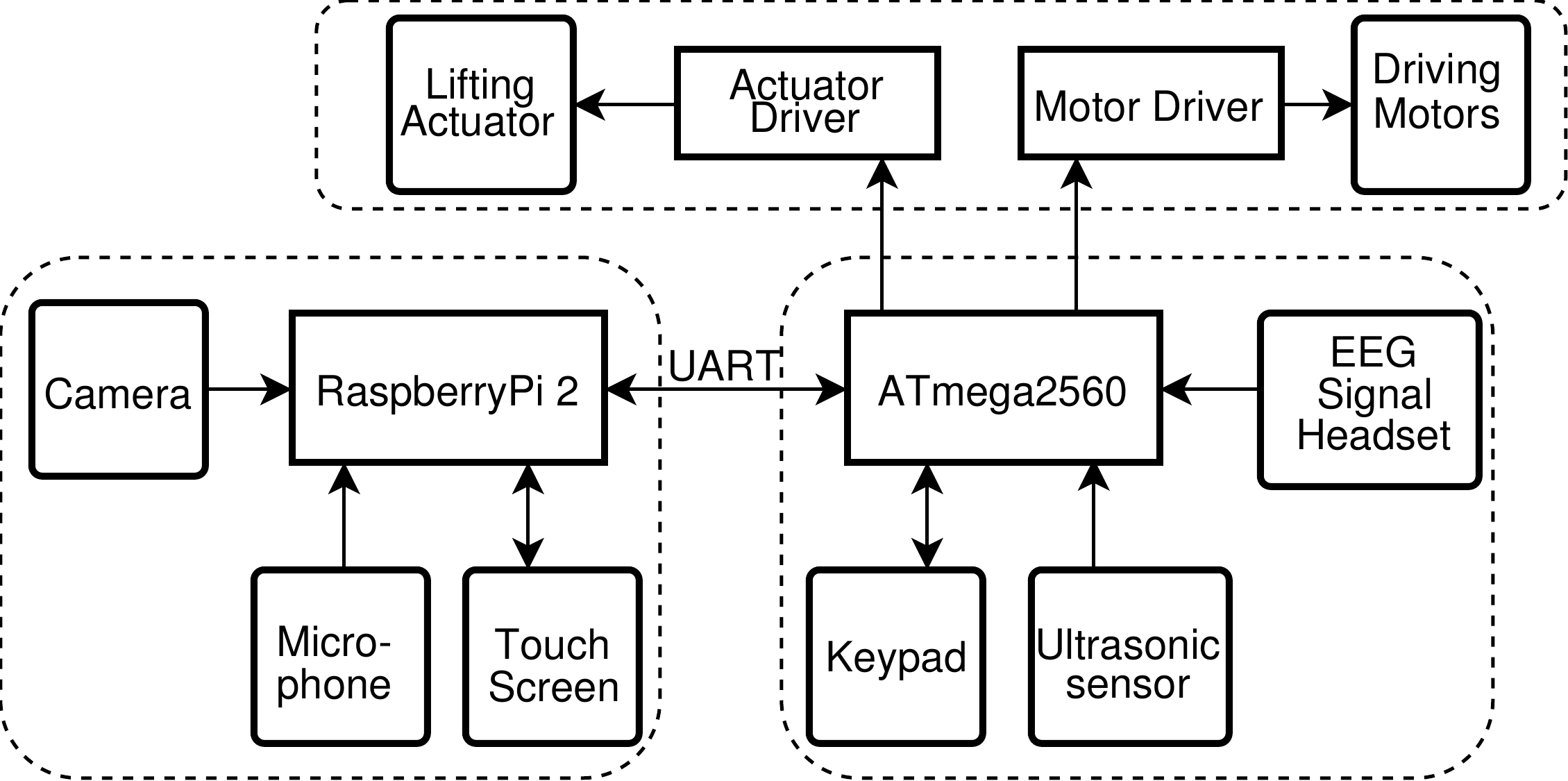}
  \caption{Embedded Architecture}
  \label{fig:embed}
\end{figure}
The control of SKALA is being divided among a low level micro-controller ATmega 2560 and a high level processor Raspberry Pi 2. The ATmega 2560 is the unit that signals all drivers, thus controlling all the motors and actuators. It is also interfaced with a hex-keypad for manual teleoperation, Ultrasonic sensors for obstacle avoidance and an EEG sensor headset for an experimental method of control. The keypad is the default mode of teleoperation that also selects/deselects other modes of control. The 3 ultrasonic sensors provide continuous range data using 3 external interrupts. The EEG headset is interfaced via UART
communication. A bidirectional serial communication is being established between ATmega 2560 and the Raspberry Pi 2, the former sending commands about switching of teleoperation modes, and receiving target motor commands. The Raspberry Pi 2 is used for high-level processing which includes voice recognition and vision-based object tracking. A 5-inch touchscreen is interfaced with Raspberry-Pi via SPI protocol
which is used for displaying live camera feed and receiving touch commands for selecting the object for tracking. For front view video feed, a Raspberry Pi Camera Module v1.3 has been used.


\subsubsection{Teleoperation via Electroencephalogram (EEG) Signals}
The vertical posture i.e. seating orientation of the user is controlled by EEG signals. \cite{rebsamen2007controlling}
The sensor used is a NeuroSky MindFlex EEG headset.\cite{katona2014evaluation}
The device is modified for being used with ATmega 2560 by externally connecting the serial transmission pin of the EEG chip TGAM1 to the Serial data receive (RX) of the ATmega. From the sampled EEG data, the useful attention and meditation parameters are extracted scaled in the range of 1 to 100 using the Arduino-Brain library. The value of meditation variable is used to monitor the state of thought of the user on the basis of which the vertical position is controlled. When the user is in state of meditation the actuator lifts him up and vice-versa. For the prevention of errors due to outliers or any artifact in data received, the data is pre-smoothened using LOESS (Local Regression) method.
\begin{figure}[!h]
\centering	
  \includegraphics[width = 0.6\linewidth]{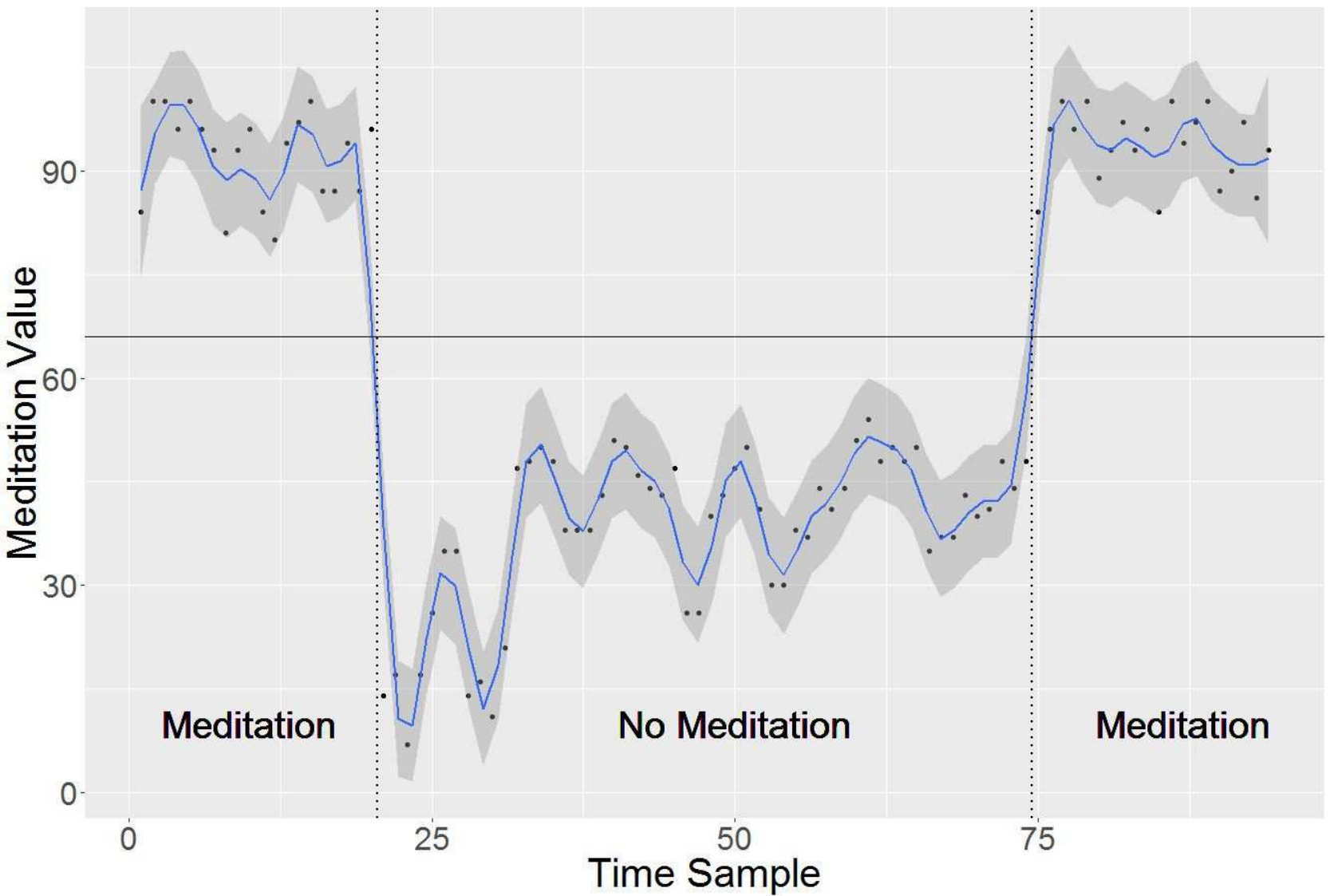}
  \caption{Data from EEG headset with regions shown}
  \label{fig:eeg}
\end{figure}

\subsubsection{Voice Teleoperation}
Voice recognition task is implemented on Raspberry Pi 2 using the open-source CMUSphinx toolkit. \cite{lamere2003cmu}
It requires a phonetic dictionary \& provides the system with a mapping of vocabulary words to sequences of phonemes. A language model defines the set of recognizable phrases to the decoder. This language model is adapted to improve the fit between the adaptation data and the model. The adaptation process improves the existing model using labeled recorded speech.
\begin{figure}[!h]
\centering	
  \includegraphics[width = 0.6\linewidth]{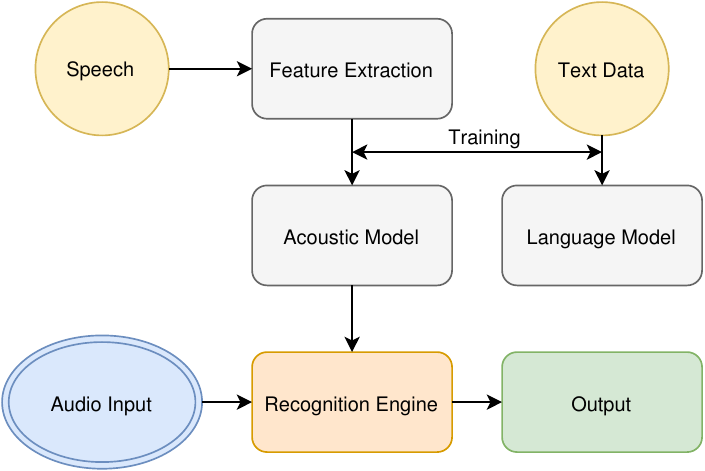}
  \caption{Overview of voice recognition API}
  \label{fig:voice}
\end{figure}


\section{Automation Architecture}
\subsection{Vision-based Object Tracking}
Computer Vision is used to track a particular object in the robot's field of view. This enables the user to traverse towards a particular destination, set by tapping once on the touchscreen display. The touchscreen is calibrated to output the camera feed with proper offset. As soon as a touch is registered, all the corner points in the current frame are detected. Then, the corner point closest to the touch coordinates is found. This point is tracked in subsequent frames using forward-backward Lucas-Kanade optical flow \cite{Lucas:1981:IIR:1623264.1623280} algorithm. The angle subtended by the object with respect to the robot is calculated using the pixel location, and appropriate motor commands are sent to the motor drivers, resulting in a continuous feedback based control system. In case, the point goes out of the frame, tracking is stopped and the program waits for the user to provide another destination.


\subsection{Obstacle Avoidance}
\begin{figure}[!h]
\centering	
\captionsetup{justification=centering}
\begin{minipage}[t]{.5\linewidth}
	\centering
  \includegraphics[width=\linewidth]{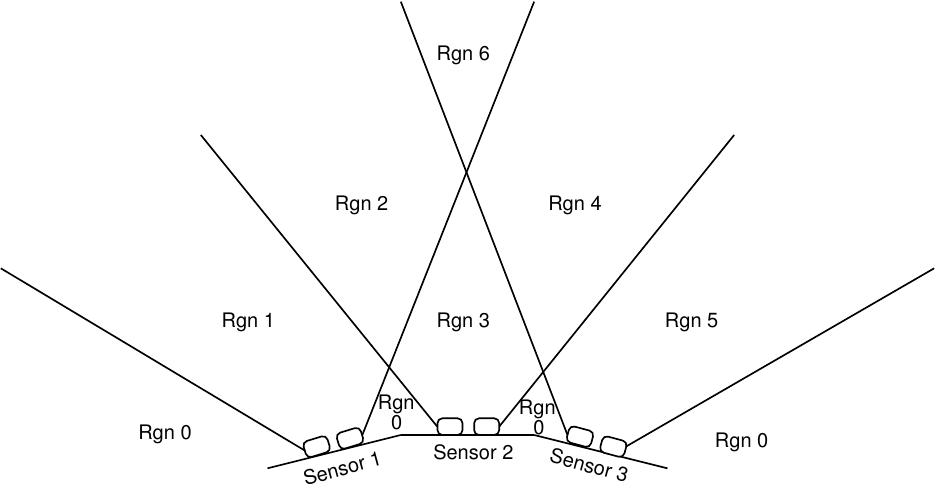}
  \captionof{figure}{Digital mapping of field of view}
  \label{fig:sonar}
\end{minipage}%
\begin{minipage}[t]{.5\linewidth}
	\centering
 \includegraphics[width=\linewidth]{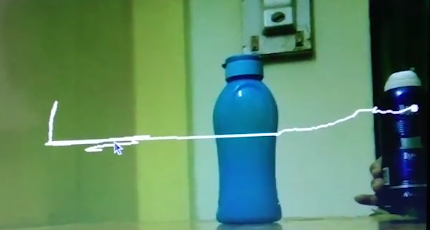}
  \captionof{figure}{Vision-based object tracking in action}
  \label{fig:object}
\end{minipage}
\end{figure}

.The ultrasonic sensors used are HC-SR04. Each of these sensors is interfaced with the ATmega controller via external interrupt mechanism . In total, three sensors are used which digitally map the robot's field of view
into 7 different regions. As shown in figure (\ref{fig:sonar}), the regions of occupancy are determined and accordingly, the motors are commanded to avoid the already occupied regions.

\section{Conclusions}
In this paper, we described the design process of semi-automated rehabilitation robot SKALA, which is capable of accommodating a person of maximum weight 120Kg with a safety factor chosen as 1.25. It is capable of running on plain ground at a maximum continuous speed of 3m/s, and has been tested to climb stairs at speed of 0.1m/s. The relevant simulations of mechanical design and logs of sensor data are also included. Basic automation methods have been incorporated into the design.
Future work with this research platform may include additional automation features such as automatic alignment with stairs and vision-based localization. The current prototpye has been fabricated with cast iron, but future work can be manufactured using more premium materials like carbon fiber, for longevity which is necessary for commercialization of the prototype.
\begin{figure}[!h]
\centering	
  \includegraphics[width = 0.7\linewidth]{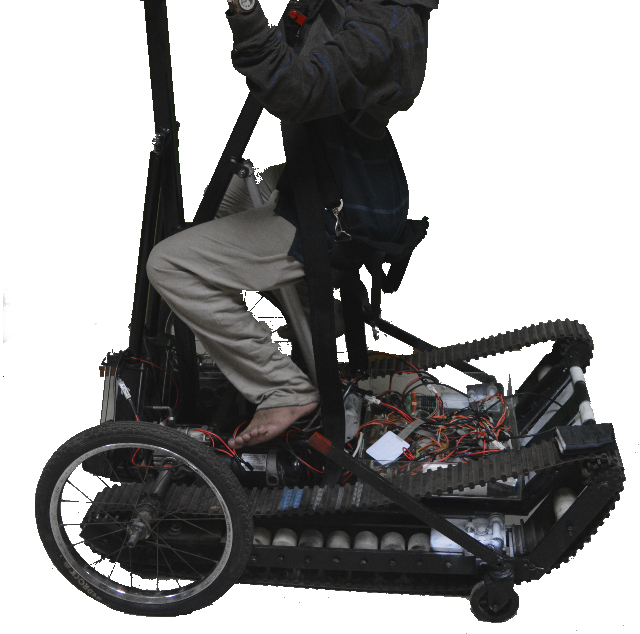}
  \caption{The Actual Prototype}
  \label{fig:skala}
\end{figure}

\section{Acknowledgments}
 We would like to express our gratitude to the management committee of the Azad Hall of Residence, IIT Kharagpur for their assistance  and  encouragement. We would also like to thank some of our colleagues Rohan Sewani, Rishu Raj, Saitanay Naribole, Nikhil Bhelave, Deepak Yadav, Shivam, Kavish, Shreyansh, Nandeesh, Devang, Kewal, Hemant and Mayank who contributed during the course of development. 

%
\bibliographystyle{unsrt}
\bibliography{sigproc}  
%
%
\end{document}